\documentclass[conference]{IEEEtran}
\IEEEoverridecommandlockouts

\usepackage{cite}
\usepackage{amsmath,amssymb,amsfonts}
\usepackage{algorithmic}
\usepackage{graphicx}
\usepackage{textcomp}
\usepackage{xcolor}

\usepackage{listings}
\usepackage{multirow}
\usepackage{array}
\usepackage{booktabs}
\usepackage{bm}
\usepackage{subcaption}
\usepackage{makecell}

\usepackage{tikz}
\usepackage{textcomp}

\def\BibTeX{{\rm B\kern-.05em{\sc i\kern-.025em b}\kern-.08em
    T\kern-.1667em\lower.7ex\hbox{E}\kern-.125emX}}
    
\newcommand\copyrighttext{%
  \footnotesize \textcopyright 2019 IEEE. Personal use of this material is permitted.
  Permission from IEEE must be obtained for all other uses, in any current or future
  media, including reprinting/republishing this material for advertising or promotional
  purposes, creating new collective works, for resale or redistribution to servers or
  lists, or reuse of any copyrighted component of this work in other works.
  DOI: 10.1109/CEC.2019.8789937}
\newcommand\copyrightnotice{%
\begin{tikzpicture}[remember picture,overlay]
\node[anchor=south,yshift=10pt] at (current page.south) {\fbox{\parbox{\dimexpr\textwidth-\fboxsep-\fboxrule\relax}{\copyrighttext}}};
\end{tikzpicture}%
}

\begin{document}
\title{Consistent Feature Construction with Constrained Genetic Programming for Experimental Physics}

\author{\IEEEauthorblockN{No\"elie Cherrier\IEEEauthorrefmark{1}\IEEEauthorrefmark{2}, Jean-Philippe Poli\IEEEauthorrefmark{1}, Maxime Defurne\IEEEauthorrefmark{2} and Franck Sabati\'e\IEEEauthorrefmark{2}}
\IEEEauthorblockA{\IEEEauthorrefmark{1}CEA, LIST, 91191, Gif-sur-Yvette cedex, France.}
\IEEEauthorblockA{\IEEEauthorrefmark{2}Irfu, CEA, Universit\'e Paris-Saclay, 91191, Gif-sur-Yvette cedex, France.}
}

\maketitle
\copyrightnotice
\begin{abstract}
A good feature representation is a determinant factor to achieve high performance for many machine learning algorithms in terms of classification. This is especially true for techniques that do not build complex internal representations of data (e.g. decision trees, in contrast to deep neural networks).
To transform the feature space, feature construction techniques build new high-level features from the original ones. 
Among these techniques, Genetic Programming is a good candidate to provide interpretable features required for data analysis in high energy physics.
Classically, original features or higher-level features based on physics first principles are used as inputs for training. However, physicists would benefit from an automatic and interpretable feature construction for the classification of particle collision events.

Our main contribution consists in combining different aspects of Genetic Programming and applying them to feature construction for experimental physics. In particular, to be applicable to physics, dimensional consistency is enforced using grammars.

Results of experiments on three physics datasets show that the constructed features can bring a significant gain to the classification accuracy. To the best of our knowledge, it is the first time a method is proposed for interpretable feature construction with units of measurement, and that experts in high-energy physics validate the overall approach as well as the interpretability of the built features.
\end{abstract}


\section{Introduction}

The performance of many supervised machine learning (ML) algorithms depends on the list of features provided with the training examples. 
Inappropriate representation of data may indeed lead to limited performance, while carefully selected features may improve the performance of the subsequent learning algorithm \cite{john_irrelevant_1994}. 
This is especially true for algorithms that have a limited ability to build an optimal internal representation of data. 
For instance, classifiers such as decision trees or more generally rule extraction algorithms represent the data as hyper-rectangles, which is not optimal in many applications but has the advantage of producing explainable outputs. 
In contrast, ``black box'' algorithms such as deep neural networks build a much more complex internal representation of data, which is often more efficient but discards any explanation for the final output \cite{gilpin_explaining_2018}.
Yet the readability of features is a requirement in several applications.
A full transparency of the model is for instance essential to its acceptance in sensitive applications such as national security, medical diagnoses or banking activities.
Additionally, the readability of features enables supervision and re-usability by experts of the field, which is needed in particular in the case of high-energy physics (HEP) experiments. 
We are interested in particular HEP experiments: using different accelerator facilities across the world, particles are collided with each other to probe their internal structure and/or search for physics beyond the Standard Model. 
Detectors are placed around the collision site for the reconstruction of the final state of the collision, i.e. the determination of the charge, mass and momentum of all particles after collision.

We focus here on distinguishing signal events from background in several HEP experiments. 
The algorithms must be trained on labeled Monte-Carlo simulated data, to be later applied to unlabeled real data from the experiments. 
Although the simulation programs are very sophisticated, they may not exactly reflect the reality (background processes may not be all included, modeled physics are not exactly the same, etc.). 
An interpretable classification model is therefore preferable to better analyze the predictions on real data. 
Based on the use of physics first principles such as momentum and energy conservation, physicists manually add high-level variables to the initial set of input features to improve the classification, whether it is based or not on machine learning algorithms. Automating the construction of such features would enable the discovery of new variables of interest that are interpretable to the experts. One would be able to build an entirely interpretable classifier, which permits a deeper performance analysis while applying the trained classifier to real unlabeled data.

We develop an algorithm derived from Genetic Programming (GP) to build automatically high-level features that are consistent with physics laws and that help the classification task. 
Next section reviews existing techniques for feature construction, and then details methods to constrain GP. 
The section afterwards presents our derived GP method adapted to experimental physics applications. 
Experiments and comparisons are conducted then. 
We finally conclude the paper and give perspectives.

\section{Related work}

\subsection{Feature construction}

Feature engineering is the general processing step that transforms the feature space to make it better suited to the learning task desired output. 
Several methods perform feature engineering indirectly, like Support Vector Machines (SVM) \cite{Vapnik1995} or Deep Neural Networks \cite{Bengio2013}.
Dimensionality reduction algorithms such as Principal Component Analysis or Linear Discriminant Analysis also build new features which are often complex non-linear combinations of the primary ones and are not interpretable.

Both feature selection and feature construction methods split into three categories. Firstly, filters rank the features independently of any classifier. Secondly, wrappers use the score obtained by a classification algorithm. Finally, embedded methods gather algorithms that perform feature construction at the same time as the optimization, such as SVM. We are interested only in the first two categories.

A review of several feature construction methods can be found in \cite{sondhi_feature_2009}.
The methods detailed in the rest of this section construct new numerical features according to a list of allowed transformations on the dataset. The new features are combinations of operators applied on the original features, and can consequently be represented by trees in which the nodes are the transformations and the leaves are the original features.
These numerical feature construction methods received increased attention in the past years. 

A number of methods use tree-based algorithms to explore the search space.
For instance, FICUS \cite{Markovitch2002} is a general framework for feature construction, generalizing several previous works in the field. It searches new numerical features using information gain in a decision tree, given an input dataset and a list of allowed transformations.
Likewise, Cognito \cite{Khurana2016} is based on a transformation tree aiming at recommending a series of transformations. The tree is explored depending on the current performances obtained within the branches. 
In \cite{Khurana2017FeatureEF}, the authors continue this work using reinforcement learning to learn the optimal exploration strategy. 
However, this last method requires to train with several datasets before application of the optimal strategy on the desired dataset.

Another class of feature construction methods gathers evolutionary methods.
Although Particle Swarm Optimization (PSO) is already widely used in feature selection, few works use PSO for feature construction. Reference \cite{xue_pso_2013} associates one particle with one constructed feature, the dimensionality being the number of original features. 
However, this method does not include the choice of operators to combine the features. 
Reference \cite{dai_new_2014} expands the dimensionality to include the selection of operators. 
They propose two representations: pair and array representation, to enable the evolution to optimize the operators as well as the features used. 
However, the space of features that can be constructed with these methods is limited, since only binary operators can be included. 
Moreover, the algorithm combines the features in a way that one cannot represent a constructed feature by a balanced binary tree.

Genetic Programming (GP) is a popular method among the feature construction methods. A survey on applications of GP to classification can be found in \cite{espejo_survey_2010}. 
GP is an evolutionary computation technique that evolves tree representations of computer programs \cite{koza_genetic_1992}. In the context of feature construction, this refers to a combination of several features thanks to a list of allowed operators (e.g. $+$, $-$, $\times$, $\div$). The main steps are as follows:
\begin{enumerate}
\item Start from a random initial population
\item Evaluate its individuals with some fitness function
\item Loop until termination criterion:
\begin{enumerate}
\item Perform selection
\item Perform mutation and crossover to get to the next generation of individuals
\item Evaluate new individuals
\end{enumerate}
\item Return individual with highest fitness
\end{enumerate}
In wrapper methods, the fitness of the individual representing an evolved feature set is the score of a predictive model using these features \cite{krawiec_genetic_2002}. 
In filter methods, the fitness measure can be for instance the information gain \cite{otero_genetic_2003,smith_genetic_2005}, entropy \cite{neshatian_genetic_2008}, or the Fisher criterion \cite{guo_feature_2005}. 
A comparison of wrapper and filter approaches in the case of genetic programming is drawn in \cite{neshatian_filter_2012}. 
GP based feature construction methods can construct one feature to add to the original feature set \cite{otero_genetic_2003,firpi_prediction_2005}, or several at a time \cite{krawiec_genetic_2002,tran_multiple_2016}, etc.

Finally, Grammatical Evolution (GE), very similar to GP, has been successfully applied to feature construction \cite{gavrilis_selecting_2008}.

We choose to adapt GP algorithm to our problem. Its main asset is its flexibility: one can build one or several features at a time, choose the maximum complexity of the features, etc.
Furthermore, there is abundant literature about techniques to add constraints to GP such as the respect of dimensional consistency (of physical units and of the feature dimensions). Several techniques have been developed to incorporate domain knowledge to constrain the evolution phase.

\subsection{Constrained Genetic Programming}

Our goal is to come up with high-level features that respect physics laws, in order to be interpretable and reusable by the experts. 
A review underlines the difficulty to incorporate domain knowledge in feature construction methods \cite{sondhi_feature_2009}.

In his early work in GP, Koza \cite{koza_genetic_1992} uses syntactic constraints to enforce the production of valid individuals. 
He imposes a root operator, or restrict the crossover operation according to the operator types for instance. 
Standard GP actually requires closure, namely that any individual tree can be considered as a subtree of another tree. 
To keep valid individuals during the entire evolution when constraining the construction process, one must modify the evolution functions. 
In Strongly Typed GP (STGP) \cite{montana_strongly_1995,haynes_type_1996}, the crossover and mutation operators are restricted according to the constraints put on the construction process.
The author of \cite{montana_strongly_1995} defines type labels and associates one to each original feature.
Every operator has then a list of accepted data types (e.g. vectors, matrices, angles, etc.) and a return type.

Constraints in GP can also be expressed by grammars: the technique is called Grammar-Guided Genetic Programming (GGGP) \cite{mckay_grammar-based_2010}.
In GGGP, an individual is a tree derived from the grammar, called a derivation tree, from which one can infer a GP-regular expression tree. 
The evolution uses the same operators than in standard GP, but applies on the derivation trees instead of the expression tree. 
However, the crossover and mutation operators are constrained so that the offspring respects the grammar.

Reference \cite{gruau_advances_1996} states that Context-Free Grammar (CFG) reduce the search space by expressing syntactic constraints on admissible individuals.
Industrial applications begin to appear at this point. 
In \cite{keijzer_dimensionally_1999}, the authors perform a multi-objective evolution to favor dimensionally consistent solutions, but non-valid individuals are still present in the population. 
In \cite{ratle_grammar-guided_2001}, a CFG constrains the evolution to produce only individuals representing dimensionally valid expressions, to discover empirical physics laws from numerical experiments.
Both applications modify the initialization procedure, which can raise an error if there is not a terminal (i.e. leave of the tree, which can be either an original feature or a constant) for each type.

Another way of using grammars in the evolution is linear GGGP. 
The individuals are represented as strings of integers, which encode a derivation sequence from the grammar. 
The derivation leads to a derivation tree and then to a syntax tree. 
The evolution is performed on the string of integers with a genetic algorithm. 
The most widely used linear GGGP is Grammatical Evolution (GE) \cite{oneill_grammatical_2001}. 
Several articles use GE for feature construction \cite{gavrilis_selecting_2008,miquilini_enhancing_2016,yazdani_mbcgp-fe_2017}.
They define a simple grammar including the operators (e.g. $+$, $-$, $\times$, $\div$) used in the construction. 
However, they do not take advantage of the potential of grammars to enforce constraints on the shape of the tree or on the validity of the use of certain operators on some variables.
Their grammar is called a universal grammar.

For physics applications, the domain knowledge involves in particular dimensionality analysis, to ensure that the constructed features are dimensionally consistent.
For this purpose, dimensional analysis can be expressed as a CFG, as proven in \cite{ratle_grammar-guided_2001}. The authors of \cite{schmidt_distilling_2009} use symbolic regression, containing a GP algorithm, to extract physics laws from experimental data. However, they ensure the dimensional consistency of the built expressions the well-chosen scalar parameters without the use of grammars.

Finally, a number of methods use Probabilistic Model-Building Algorithms (PMBA) \cite{kacprzyk_survey_2006}. 
The idea is to maintain a distribution estimation over the search space. 
An initial probability distribution is defined on the grammar, and then the fitnesses of the individuals are used to update these probabilities \cite{ratle_avoiding_2006}. 
Their objective is to guide the search.

To deal with HEP applications, we propose to adapt the grammar of \cite{ratle_grammar-guided_2001} to handle dimensional consistency.

\section{Genetic Programming adapted to experimental physics}

In this section, we detail how to take advantage of the methods listed in the previous part to come up with a feature construction algorithm adapted to HEP applications. 
Firstly, we focus on the definition of a grammar suitable for dimensional consistency of physical quantities, and the coupled probability distribution. 
Then we detail the evolution methods for our Probabilistic GGGP algorithm.

\subsection{Grammar definition}

In experimental physics, particle detectors measure numerical values that are then reconstructed into variables such as momenta ($MeV/c$), energies ($MeV$), angles ($rad$), durations ($ns$), lengths ($cm$), etc. 
Features constructed by a GP algorithm should be dimensionally consistent to be independent of the system of measurement of the base features, and to ensure their interpretability.
For instance, combinations such as adding or subtracting energies and lengths are forbidden.

We define a CFG similar to the one in \cite{ratle_grammar-guided_2001}, while adding additional restrictions on the types that can be mixed. 
For instance, we do not allow the creation of any combination of energies and lengths, or angles and durations, etc., since these combinations are less common in the HEP domain.
Figure~\ref{fig:grammar} shows an example of the grammar used for one particular dataset in our study.
For the sake of comprehension, this grammar is a simplified version of the one used in the experiments.
The GGGP can be seen in our case as a STGP, the types being the ones described in the grammar: a fixed number of types such as Energy, Angle, Distance, Float, etc.
In our study, the operations used are the same between the different datasets, and are chosen among standard operators used in HEP.

\begin{figure}
\centering
\begin{lstlisting}[
    basicstyle=\footnotesize,
]
<start> ::= <E> | <A> | <F>
        
<E> ::= <E> + <E> | <E> - <E> | <E> * <F> 
  | <E> / <F> | Sqrt(<E2>) | <termE>

<A> ::= <A> + <A> | <A> - <A> | <A> * <A> 
  | Acos(<F>) | Asin(<F>) | Atan(<F>) 
  | <termA>

<F> ::= <F> + <F> | <F> - <F> | <F> * <F> 
  | <E> / <E> | <A> / <A> | <F> / <F> 
  | Cos(<A>) | Sin(<A>) | Tan(<A>) 
  | <termF>

<E2> ::= <E2> + <E2> | <E2> - <E2> 
  | <E2> * <F> | <E2> / <F> 
  | Square(<E>) | <termE2>
\end{lstlisting}
\caption{Production rules of the grammar used for Higgs dataset (E: Energy; E2: squared Energy; A: Angle; F: Float). \lstinline{<termX>} means either a constant or base feature of type X.}
\label{fig:grammar}
\end{figure}

\subsection{Transition matrix definition}

We go further in the guidance of the evolution process to obtain physically interpretable and consistent features.
HEP theory indeed involves a number of typical formulas. 
For instance, one can study the conservation of energy and momentum in a collision of particles to infer the intermediate states, or reconstruct the collision vertex with geometrical operations. 
An automated classifier could benefit from such useful features.
An idea is then to guide the construction of the trees through the choice of a probability distribution over the grammar, to favor the construction of formulas that are similar to those used in HEP.
We choose not to update the probabilities during the evolution contrary to the strategy in \cite{ratle_avoiding_2006}, to avoid converging too quickly and to guide the search to build physically sound features by following recurrent patterns in physics formulas.

More specifically, we put constraints on the transitions between the operators. 
For instance, a square root in physics is very often followed by a sum of squares.
In this way, we also forbid a square operator followed by a square root and conversely, to simplify the trees. 
This particular constraint could actually appear in the grammar itself, but it would lead to a more complex and less readable grammar.
A probability transition matrix between a few operators is shown in Table~\ref{tab:probas}.
To this matrix one must add the initial probability distribution $P_{init}$ for the choice of the first operator.
For instance, a square root of a squared energy is much more likely to appear in HEP formulas than the division of a time by a cosine.
The probabilities are chosen manually to emulate the frequencies of operators in HEP formulas.
In the experiments, we compare the chosen probabilities to the uniform case where all operations can be selected with the same probability.

\begin{table}
\centering
\caption{Snippet of the transition matrix: the probability to choose an operator among the rows according to the previous operator in the columns.}
\label{tab:probas}
\begin{tabular}{|cc|c|c|c|c|c|c|}
\cline{3-8}
\multicolumn{2}{c|}{} & \multicolumn{3}{c|}{Energy} & \multicolumn{3}{c|}{$\textrm{Energy}^2$} \\
\multicolumn{2}{c|}{} & $+$ & $-$ & $\sqrt{}$ & $+$ & $-$ & $^2$ \\
\hline
\multirow{5}{*}{\rotatebox[origin=c]{90}{Energy}} & $+$ & 0.1 & 0.225 & & & & 0.8 \\
 & $-$ & 0.1 & 0.225 & & & & 0.1 \\
 & $\times$ & 0.1 & 0.25 & & & & 0.07 \\
 & $\div$ & 0.1 & 0.2 & & & & 0.03 \\
 & $\sqrt{}$ & 0.6 & 0.1 & & & & 0 \\
\hline
\multirow{4}{*}{\rotatebox[origin=c]{90}{$\textrm{Energy}^2$}} & $+$ & & & 0.7 & 0.4 & 0.2 &  \\
 & $-$ & & & 0.25 & 0.15 & 0.07 &  \\
 & $\times$ & & & 0.05 & 0.05 & 0.03 &  \\
 & $^2$ & & & 0 & 0.4 & 0.7 &  \\
\hline
\end{tabular}
\end{table}

\subsection{Evolution methods}

\begin{figure}
\begin{subfigure}[b]{0.6\linewidth}
\centering
\includegraphics[width=\linewidth]{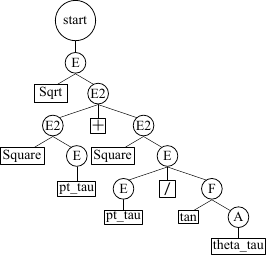}
\subcaption{}
\end{subfigure}
\hfill
\begin{subfigure}[b]{0.35\linewidth}
\centering
\includegraphics[width=\linewidth]{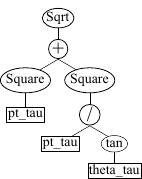}
\subcaption{}
\end{subfigure}
\caption{(a) Example of derivation tree and (b) Example of expression tree.}
\label{fig:derivation_expression_trees}
\end{figure}

From the grammars and transition probabilities, one can derive a derivation tree and then the standard GP expression tree.
Examples of a derivation tree and its associated expression tree for HEP are displayed on Figure~\ref{fig:derivation_expression_trees}.

We actually combine both techniques of GE and GP by taking advantage of standard GP evolution methods applied to expression trees, while adding a few methods replicating the effects of those in GE.
For single feature construction, an individual is one expression tree.
For multiple feature construction, an individual is a fixed-length list of $n>1$ expression trees.

An initial population is firstly evaluated, then evolved according to a $(\mu + \lambda)$-strategy. For each individual in the population, mutation can be applied with a probability $P_{mutation}$ and crossover with a probability $P_{crossover}$. 
The offspring is then evaluated, and the selection is performed over the whole parent population and the offspring, which differs from the standard $(\mu + \lambda)$-strategy but has the advantage of keeping in the population some efficient features.

In the following paragraphs, we detail the evolution methods used in our work for both single and multiple feature construction.
In the case of single feature construction, an individual is a single tree. For multiple feature construction, an individual is represented by a list of $n$ trees.

\subsubsection{Initialization}

Whether for constructing one tree (i.e. one feature) or a fixed number $n>1$ of trees (i.e. $n$ features), the trees are built the same way: they are generated using the ramped half and half initialization \cite{koza_genetic_1992}. 
The process of choosing the operators is however altered to respect the grammars.
A type $t$ and the first operator (returning type $t$) are selected under the $P_{init}$ distribution.
Then, while the condition on current depth is not satisfied, the tree keeps growing.
The possible operators are selected according to the grammar.
The transition matrix then defines a distribution probability among the remaining possible operators according to the parent node.
Finally, the leaves of the tree are randomly chosen among the set of base features of the proper type and constants.
After the population is initialized, the evolution process begins with a series of mutations, crossovers, evaluations and selections.

\subsubsection{Mutation and crossover}

Both crossover and mutation apply on the expression trees themselves, whether several features are constructed at the same time or only one. 
Regardless of the applied transformation, a node is picked in the tree following a probability distribution that depends on the previous scores achieved by the node in its tree. 
At each generation, a weight is indeed assigned to each node of every tree in the population. 
Each node contains the information of its best gain on accuracy obtained within the tree, even if other branches of the tree changed. The list of the gains is then inversed and normalized to get the weights used to select the node during mutation/crossover. In this way, the weights sum to 1, with a small shift to let a chance for the best nodes to be mutated anyway.
When mutating or mating trees, the weights are used to select the less efficient nodes with higher probability.

The mutation method is randomly picked among three existing techniques. Each of these techniques is modified to be compliant with the grammars:
\begin{itemize}
\item Uniform mutation: a node is selected in the tree, then the subtree is entirely regenerated from that node while making sure that the dimensional consistency is still respected in particular at the root.
\item Node replacement: a node is selected and replaced with any dimensionally compliant node.
\item Insertion: from a selected node, a new subtree is inserted that have the original subtree(s) of the mutated node as child nodes. The inserted subtree is generated so that the grammar is also respected at the connections with the original tree.
\end{itemize}
The transition matrix is used each time a new tree or subtree needs to be generated to support the interpretability.

The crossover operation is the standard GP one-point crossover, assuming that the exchange of the two subtrees is compatible with the grammar, i.e. that the two roots of the subtrees are of the same type.

In multiple feature construction, mutation is applied on one or several features of the selected individual. Crossover is the standard genetic algorithm one-point crossover, with no modification of the trees themselves.

The individuals created through these operations must then be evaluated and selected with the parent population to form the next generation.

\subsubsection{Evaluation and selection}

A repeated tournament selection among three individuals constitutes the next generation population. The tournament size of three was chosen experimentally as a good compromise between randomness and quick convergence.

To evaluate an individual, expression trees are converted to numerical features, by computing the function they represent.
In our implementation, a division by zero would create a missing value.
If the feature is not well constructed and the whole column is invalid, then the individual is given a fitness score of zero.
Two main approaches exist to evaluate individuals in feature construction. 
Wrapper methods evaluate the score (e.g. the accuracy) of a classifier trained on the set containing the constructed feature(s).
In this case, the score is obtained through a k-fold cross-validation on the training set.
Filter methods use ranking measures that are independent of any predictor.

In the experiments, we compare among others the performances obtained with several classifiers and with one filter fitness measure.

\section{Experiments}

In this section, we present the datasets used for the experiments, the experimental settings, and then several comparisons of methods and settings. Finally, we discuss the results.

\subsection{Datasets}

We use three datasets taken from three different and unrelated experiments.
The Higgs dataset is unbalanced, while the two other datasets are more balanced.

\subsubsection{Higgs dataset}

ATLAS is a particle physics experiment at CERN.
In 2014, the collaboration opened a Kaggle challenge\footnote{www.kaggle.com/c/higgs-boson} with the objective to automatically detect the Higgs boson decaying into two \textbf{$\tau$}-leptons in Monte-Carlo simulated data. 
The data has then been publicly released on the Open Data platform of CERN\footnote{http://opendata.cern.ch/record/328}.
The dataset \cite{adam-bourdarios_learning_2014} consists in more than 800000 simulated events, including about 280000 signal (positive) instances, the rest being background. 
17 primitive features per event are available, including notably the transverse momentum $p_T$ (in the xy-plane), $\phi$ angle (between the x and y-axes) and $\theta$ angle (between the z-axis and the xy-plane) of several particles. 
Besides, we remove 13 high-level features also provided in the dataset since our objective is to automatically make new features from the basic ones.

\subsubsection{DVCS dataset}

The experimental Hall B of the Jefferson Laboratory conducts experiments on nuclear and particle physics. 
The CLAS12 collaboration notably studies Deeply Virtual Compton Scattering (DVCS), to probe the internal structure of the proton. 
The objective of this dataset is to discriminate between two event types: DVCS and $\pi^0$-production  acting as background for the DVCS study.
The dataset consists in Monte-Carlo simulated data.
We train on about 25000 examples including 15000 DVCS (signal) events and the rest pion production events. The events are selected so that exactly three particles have been detected.
Features of these three particles include their 3-momentum, vertex position, charge, deposit energy, etc. 
A total of 36 features is available to train the model.

\subsubsection{$\bm{\tau \rightarrow \mu + \mu + \mu}$ dataset ($\bm{\tau\mu^3}$)}

The Large Hadron Collider beauty experiment (LHCb) searches for new physics at CERN. 
A Kaggle challenge\footnote{www.kaggle.com/c/flavours-of-physics} took place in 2015 to design a model able to find the decay $\tau^- \rightarrow \mu^+ \mu^- \mu^-$ among background events. 
The particularity of this dataset is that signal events are simulated data, since this decay is not supposed to happen according to the Standard Model \cite{lhcb_collaboration_search_2015}. Observation of this decay would mean a violation of the latter and consequently a sign of new physics.
Background events are taken from real experiments at LHCb.
The total dataset includes more than 67000 instances of which 40000 are signal events.
46 features describe each event, including features on the three muons and the reconstructed $\tau$-lepton. 
Features of the latter are therefore removed since they are derived from the three muons.
It should be noted that the dataset originally comes with an agreement dataset and a correlation dataset, to ensure a few constraints are respected for the further physics analysis. 
We do not use these side datasets in our work.

\subsection{Experimental settings}

We systematically present the results of our experiments by showing the mean and standard deviation over at least 20 independent runs for each method and dataset.
The settings are as follows: we evolve a population of 500 individuals over 150 generations. 
The individuals can consist in one or several trees, i.e. features.
We set the probability of mutation and crossover both to 0.6, which means that individuals can undergo both mutation and crossover during the same generation.
The trainings are done on two thirds of the datasets, through a 3-fold cross-validation on the training sets. The gains in accuracy are computed on the test sets, i.e. one third of the datasets.

\subsection{Performance comparisons}

\subsubsection{Comparison of methods}

In this paragraph, the fitness function is the mean accuracy of a XGBoost classifier over a 3-fold cross-validation on the training sets.

We first compare the performances on the three datasets of three methods constructing one feature: simple GP (without any added constraint), GGGP with the grammar adapted to the dataset, and PGGGP with the same adapted grammar and an empirically designed transition matrix.
Table~\ref{tab:comparisons} shows the gains in accuracy for the different methods and datasets, as well as the p-values from a Welch’s t-test to compare PGGGP to simple GP. The baseline accuracy is obtained by training XGBoost on the base feature set. In addition, we also present in Table~\ref{tab:comparisons} the results and statistical test of the PSO algorithm from \cite{dai_new_2014} (array representation) applied on the Higgs dataset.

\begin{table}
\centering
\caption{Gains in accuracy (in \%) obtained on three test datasets constructing one feature. First p-value compares the best of PGGGP and GGGP to simple GP. Second p-value compares the best of PGGGP and GGGP to PSO when applicable.}
\label{tab:comparisons}
\begin{tabular}{c|c|c|c}
\hline
 & Higgs & DVCS & $\tau\mu^3$ \\
\hline
baseline & $75.46$ & $66.95$ & $82.24$ \\
\hline
PSO & $0.54 \pm 0.15$ & & \\
simple GP & $1.92 \pm 0.11 $ & $0.52 \pm 0.27$ & $\bm{0.58 \pm 0.36}$ \\
GGGP & $\bm{2.23 \pm 0.68}$ & $0.86 \pm 0.27$ & $0.43 \pm 0.28$ \\
PGGGP & $2.10 \pm 0.72$ & $\bm{0.87 \pm 0.40}$ & $0.54 \pm 0.37$ \\
p-value 1 & $10^{-17}$ & & \\
p-value 2 & 0.010 & $10^{-4}$ & 0.661 \\
\hline
\end{tabular}
\end{table}

The performances vary with the datasets used.
For the DVCS dataset, the PGGGP improves the score of the XGBoost classifier by 0.87\%, which is comparable to the method without probabilities and significantly better than the GP with no constraints at all.
For the $\tau\mu^3$ dataset, the PGGGP algorithm achieves performance similar to the unconstrained GP but slightly better than the GP without probabilities, with a gain of 0.54\%.
However, for the Higgs dataset, the PGGGP method is overcome by the GGGP without transition matrix but the constrained methods are still significantly better than the simple GP. Moreover, the three GP methods significantly outperform the PSO algorithm.

\begin{table}
\centering
\caption{Gains obtained on the Higgs dataset constructing from one to six features. The p-values are not significant here hence not shown.}
\label{tab:multiple_higgs}
\begin{tabular}{cccc}
\hline
 & 1 feature & 2 features & 3 features\\
\hline
GGGP & $\bm{2.23 \pm 0.68}$ & $2.24 \pm 0.18$ & $\bm{2.61 \pm 0.16}$ \\
PGGGP & $2.10 \pm 0.72$ & $\bm{2.31\pm 0.15}$ & $2.53 \pm 0.25$ \\
\hline
\hline
 & 4 features & 5 features & 6 features\\
\hline
GGGP & $2.92\pm 0.28$ & $3.16 \pm 0.29$ & $3.24 \pm 0.31$ \\
PGGGP & $2.92\pm 0.27$ & $3.16 \pm 0.26$ & $3.24 \pm 0.25$ \\
\hline
\end{tabular}
\end{table}

To go further in the study and try to understand the discrepancies in the performances of the different methods on the Higgs dataset, we perform several experiments constructing $n>1$ features.
Table~\ref{tab:multiple_higgs} shows the gains in accuracy obtained on the Higgs dataset while constructing one to six features comparing our GGGP algorithm with or without the use of the transition matrix.
Although the GGGP algorithm performs the best when constructing a low number of features, the gap between the GGGP and the PGGGP algorithms reduces as the number of features increases, and the PGGGP algorithm overcomes the GGGP algorithm when constructing six features.

\subsubsection{Comparison of fitness functions}

In previous paragraphs, the experiments were conducted with the score of XGBoost as the fitness function.
We assess the utility of our PGGGP also when using other fitness functions.
In the experiments, we compare the performances obtained with different classifiers: a single decision tree (DT), the k-nearest neighbors algorithm (KNN) and the naive bayes algorithm (NB) as wrapper evaluators, and with the entropy of \cite{neshatian_filter_2012} as a filter fitness measure.

\begin{table}
\centering
\caption{Gains on accuracies obtained on the Higgs dataset constructing one feature with different fitness functions, with the associated p-values comparing simple GP to PGGGP.}
\label{tab:fitness_fcts}
\begin{tabular}{c|ccc|c}
\hline
 & baseline & simple GP & PGGGP & p-value\\
\hline
DT & 66.92 & $2.74 \pm 0.19$ & $\bm{2.96 \pm 0.58}$ & 0.035\\
KNN & 71.18 & $1.09 \pm 0.35$ & $\bm{1.52 \pm 0.66}$ & 0.001\\
NB & 62.95 & $4.11 \pm 8.55$ & $\bm{6.49 \pm 1.51}$ & 0.119\\
entropy & 75.46 & $0.07 \pm 0.12$ & $\bm{0.11 \pm 0.10}$ & 0.181\\
\hline
\end{tabular}
\end{table}

Table~\ref{tab:fitness_fcts} compares the accuracies obtained with these different fitness measures and gives the p-values from a Welch's t-test. 
In the case of the entropy measure, the scores are computed with a XGBoost classifier trained on the datasets created by the evolutionary algorithm.
Using any fitness function, our probability-aided GGGP algorithm improves the base score compared to the simple unconstrained GP algorithm. The improvement is significant for the DT and the KNN.
These results show the robustness of our algorithm to several fitness functions.

\subsection{Discussion}

We observe in Table~\ref{tab:comparisons} that the method with probabilities reaches lower performances than the GGGP for the Higgs dataset, in particular when constructing one feature only.
Actually, the GGGP comes up with a feature that achieves a gain of $+3.82\%$:
{\fontsize{8.5}{9.5} \selectfont
\begin{equation}
\frac{\sqrt{p_t^{lep}\left( missingtE + p_t^{lep} + p_t^{\tau} \right)} + \sqrt{m_{H^0}^2 + missingtE^2 + {p_t^{\tau}}^2}}{\left(\cos\left( \phi^{lep} - \phi^{\tau} \right) + 2\right)^2 \cos^4\left( \theta^{lep} - \theta^{\tau} \right)}
\label{eq:big_feat}
\end{equation}
}
In HEP, physicists may perform cuts on hand-crafted high-level features to isolate signal events from background events, thus building a rule of inference. 
The feature \eqref{eq:big_feat} above is mathematically correct, but does not have an intuitive meaning to the physicists. However, it is independent of the system of measurement and can therefore be used to replace the usual features of the physicists.
Although the ratio is not familiar to the physicists, the elementary components can be interpretable.
The numerator indeed resembles a transverse energy/momentum balance.
In the denominator, the elements are more interpretable. 
As expected, the $\theta$ and $\phi$ angles are not mixed since they are defined in orthogonal planes. 
From energy/momentun conservation, a strong correlation exists between the direction of the lepton and the $\tau$-lepton. 
Therefore, comparing the two angles of the two particles makes completely sense. Because the angles are defined within $]-\pi;\pi]$, the comparison is much easier with the cosine function.

\begin{figure}
\centering
\includegraphics[width=0.95\linewidth]{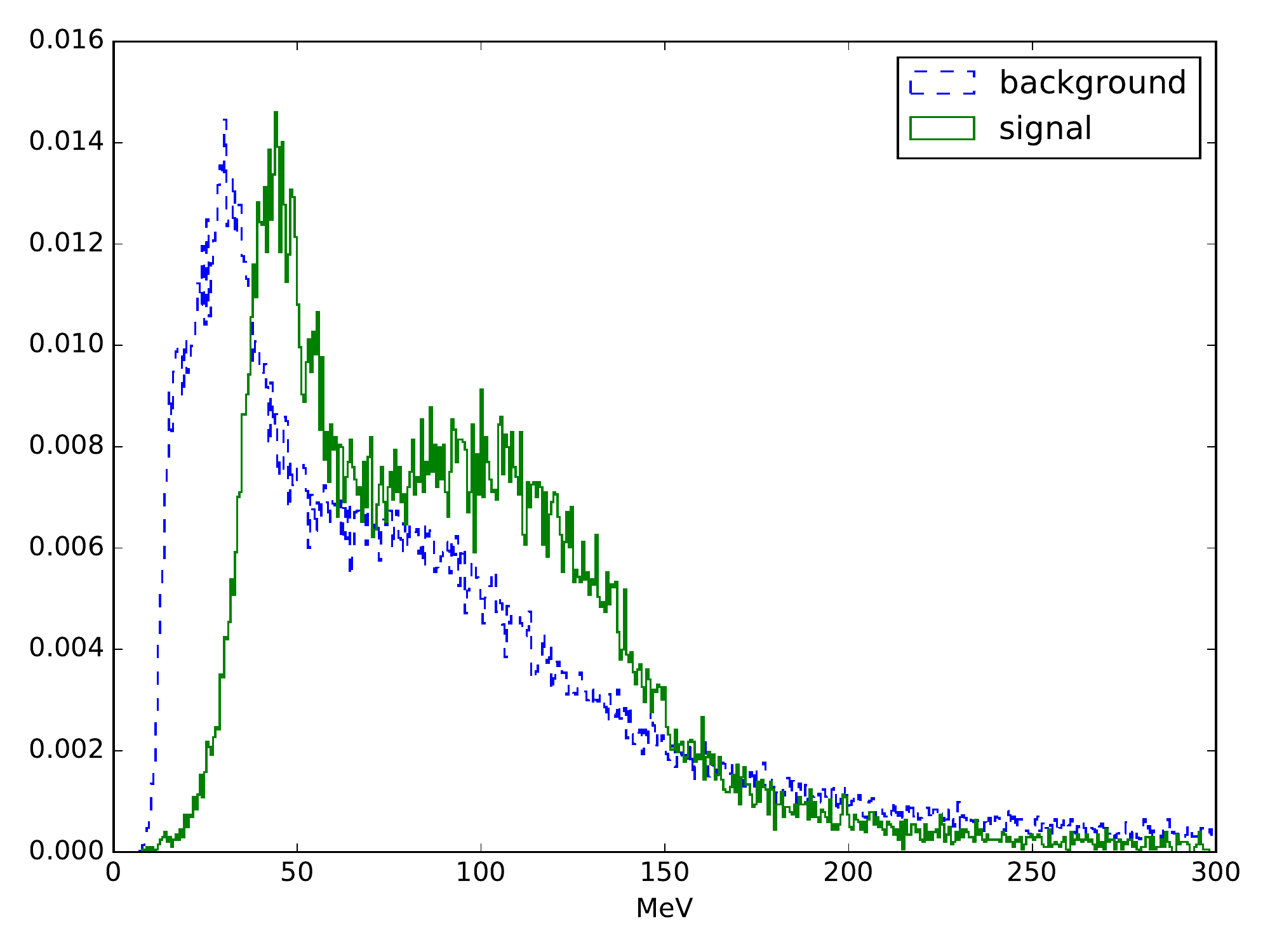}
\caption{Histogram of the feature constructed by the GGGP algorithm.}
\label{fig:distrib}
\end{figure}

Figure~\ref{fig:distrib} shows the histogram of the feature, visually proving its discriminating power.
The physicists widely use the invariant mass as a high-level feature to improve the classification.
The invariant mass is a 4-momentum balance between two particles, here the tau and the lepton.
Adding this invariant mass to the original dataset increases the score by $2.91\%$.
With the GGGP, we overcome the invariant mass and provide a better performing feature which is independent of the system of units, adding $3.82\%$ to the accuracy.

However, the GGGP with transition matrix does not achieve this level of performance, probably because the ratio between an energy and a series of cosines is set to be very unlikely.
When constructing several features at a time, the performances of the two methods become in fact similar (see Table~\ref{tab:multiple_higgs}).
The GGGP without probabilities is then confronted to a bigger search space without guidance, whereas the GGGP with transition matrix gets more space to construct interpretable and performing features all at once.
For instance, the GGGP using the transition matrix builds five features achieving a gain of $+3.71\%$ on the accuracy:
\begin{equation}
\label{eq:multifeat1}
\cos \left(\phi^{lep} - \phi^{\tau}\right)
\end{equation}
\begin{equation}
\label{eq:multifeat2}
\cos\left(\theta^{lep} - \theta^{\tau}\right)
\end{equation}
\begin{equation}
\label{eq:multifeat3}
\cos\left(\phi^{missing} - \phi^{lep}\right)
\end{equation}
\begin{equation}
\label{eq:multifeat4}
p_{T}^{leading} \sum p_{T}^{jets} - \left(E_{T}^{missing} + p_{T}^{lep}\right)^2
\end{equation}
\begin{equation}
\label{eq:multifeat5}
m_{H^0}^2 + \left(p_{T}^{lep} + p_{T}^{\tau}\right)^2
\end{equation}
The first and second features \eqref{eq:multifeat1} \eqref{eq:multifeat2} are exact elementary components of the single feature constructed by the GGGP. The third \eqref{eq:multifeat3} is another geometrical combination of the angles of the missing transverse energy (most likely corresponding to a neutrino) and the lepton, which both come from the same $\tau$ lepton.
It should be noted that the $\theta$ and $\phi$ angles are almost never mixed in the output features, proving that the GP algorithm inferred that these two angles belong to two different planes.
The two last features \eqref{eq:multifeat4} \eqref{eq:multifeat5} are energy balances in the transverse space.
These five constructed features are more interpretable features than the one from the simple GGGP, and are also independent of the system of measurement and therefore reusable in HEP analyses.

Finally, it seems that forcing the algorithm to build interpretable features is not beneficial when constructing only one feature. Therefore, the GGGP manages to compress the information of several interpretable features into one, whereas the GGGP aided by the transition matrix is less efficient. Actually, the best features built by the PGGGP in the context of single feature construction are very complex and exploit the maximum allowed depth of the trees. However, having big trees impairs the interpretability of the associated features. In this context, the preference of PGGGP over GGGP for the interpretability of the features is not obvious.

\begin{figure}
\centering
\includegraphics[width=0.95\linewidth]{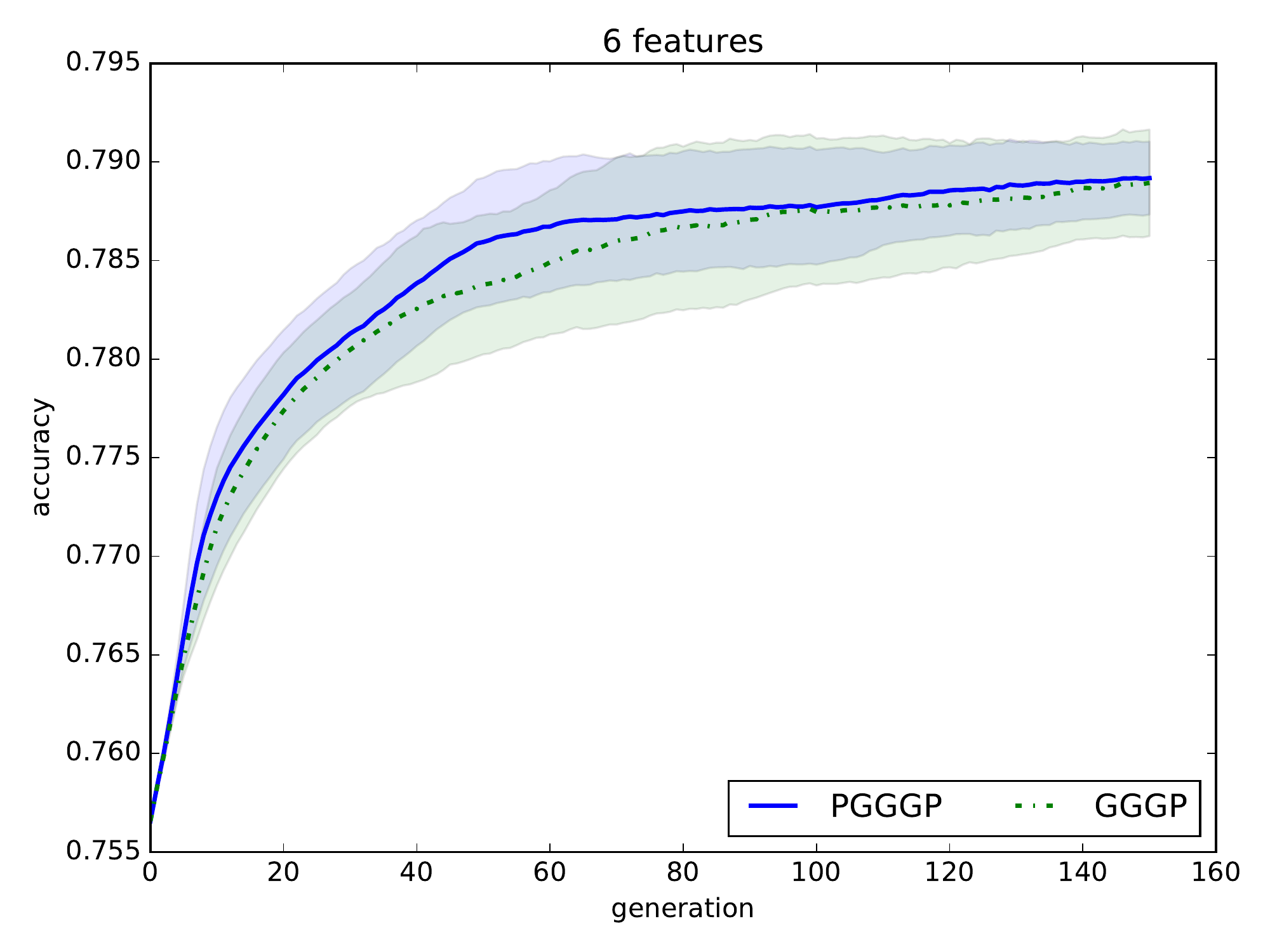}
\caption{Learning curve for constructing 6 features on Higgs dataset. The mean and standard deviation of the best fitness of the population are shown at each generation.}
\label{fig:learning_curves}
\end{figure}

However, the transition matrix seems to help the GGGP algorithm to converge faster (see Figure~\ref{fig:learning_curves}).

On the DVCS dataset, probabilities also seem to help the convergence and performance of the GGGP algorithm. However, the statistical tests to prove the difference between GGGP and PGGGP results are never significant regardless of the dataset. PGGGP and GGGP algorithms are actually very similar in their conception, and only differ by the choice of the probabilities to construct the trees. It is therefore normal not to observe any significant variation in the distributions of the results. Only slight improvements can be observed.

Finally, the performance gains on the $\tau\mu^3$ dataset are lower than those on the other two datasets.
This can be explained by the design of the dataset itself: providing many high-level variables (we partly discarded) and missing basic features such as the $\phi$-angle essential for our experiment, this dataset might not be suited for our study.
With the features provided in the dataset, more complex non-linear and therefore non-interpretable combinations are probably more efficient to increase the accuracy of a classifier.
This can also explain the better score of the simple GP, which is not constrained and then allowed to apply complex operations such as trigonometric functions on all variables.
The features built by our GGGP, with or without probabilities, are already quite complex: the most performing individuals have a height $h \geq 6$.
However, we still manage to increase the accuracy score by 1.48\% constructing one feature with PGGGP.

Finally, one of our objectives being the interpretability of the built features, it is difficult to provide an accurate comparison with other methods that do not focus on interpretability or consistency of features, or that have different structure for built features. Therefore, the comparison can be only on the numerical scores.
Within evolutionary methods other than GP, we tested a PSO algorithm, which performs poorly on our datasets. We also considered tree-based algorithms such as Cognito or FICUS, but the lack of open source code and implementation details in the papers prevented us to draw a fair comparison.
Besides, the datasets used in other studies do not include units of measurement. Few benchmarks with units of measurement exist, since the data are often anonymized. However, our method is useful for many real-world problems.

\section{Conclusion and perspectives}

In this paper, we have presented an adaptation of a new Grammar-Guided Genetic Programming algorithm to dimensional consistency and physics laws, aided with probability transition matrices.
Compared to a simple Genetic Programming algorithm without constraints, we have shown that this method significantly improves the accuracy score of several classifiers for two high-energy physics datasets from completely different experimental setups. 
The possibilities to build new features for the third dataset are limited due to the lack of base features.
However, for the three datasets, our interpretable GGGP-based feature construction algorithm brings a significant improvement on the classification accuracy especially for datasets with a low baseline score.
We discussed the trade-off dilemma between performances and interpretability raised by the application of the transition matrix.
The comparison between GGGP and PGGGP in gain in accuracies mainly relies on the data on which the experiment is conducted and on the number of features that are constructed. The interpretability for physics experts is definitely better with PGGGP, but the score can be better with GGGP (without probabilities) because of the need to compress information into a small number of features. Therefore, we observe that constructing several features enables the PGGGP to fill the gap with the GGGP while being more interpretable.

In future work, we plan to automatically extract the probabilities of the transition matrix from physics formulas taken from books or articles.
Another perspective would be to enforce the impact of probabilities as we go deeper in the generation of the tree: this could help building consistent elementary components while enabling more complex combinations at the root of the tree, imitating the high performing feature obtained by the GGGP on the Higgs dataset.

\section*{Acknowledgement}
We want to thank Marouen Baalouch for his precious help during the experiments, Francesco Bossu and Herv\'e Moutarde for their help understanding High-Energy Physics and the datasets, and more generally the CLAS12 collaborators for the simulation software.

\bibliographystyle{IEEEtran}
\bibliography{refs}

\end{document}